\documentclass{bmvc2k}
\usepackage{amsfonts}

\title{\hspace{0.7cm}What Face and Body Shapes Can Tell\\\hspace{0.7cm}About Height}

\addauthor{\hspace{0.7cm}Semih G\"{u}nel}{semih.gunel@epfl.ch}{1}
\addauthor{\hspace{0.7cm}Helge Rhodin}{helge.rhodin@epfl.ch}{1}
\addauthor{\hspace{0.7cm}Pascal Fua}{pascal.fua@epfl.ch}{1}

\addinstitution{ 
  CVLab\\
  EPFL,\\
  Lausanne, Switzerland
}

\runninghead{G\"{u}nel et al.}{What Face and Body Shapes Can Tell About Height}

\begin{document}

\maketitle

\definecolor{olive}{RGB}{50,150,50}

\newif\ifdraft
\drafttrue

\ifdraft
 \newcommand{\PF}[1]{{\color{red}{\bf pf: #1}}}
 \newcommand{\pf}[1]{{\color{red} #1}}
 \newcommand{\HR}[1]{{\color{blue}{\bf hr: #1}}}
 \newcommand{\hr}[1]{{\color{blue} #1}}
 \newcommand{\VC}[1]{{\color{green}{\bf vc: #1}}}
  \newcommand{\vc}[1]{{\color{green} #1}}
 \newcommand{\ms}[1]{{\color{olive}{#1}}}
 \newcommand{\MS}[1]{{\color{olive}{\bf ms: #1}}}
 \newcommand{\JS}[1]{{\color{cyan}{\bf js: #1}}}
 \newcommand{\NEW}[1]{{\color{red}{#1}}}
 \definecolor{turquoise}{rgb}{0.19, 0.84, 0.78}
 \newcommand{\sg}[1]{{\color{turquoise}{\bf sg: #1}}}
 \newcommand{\SG}[1]{\color{turquoise}{#1}}
 
\else
 \newcommand{\PF}[1]{{\color{red}{}	
 \newcommand{\pf}[1]{ #1 }
 \newcommand{\HR}[1]{{\color{blue}{}
 \newcommand{\hr}[1]{ #1 }
 \newcommand{\VC}[1]{{\color{green}{}
 \newcommand{\ms}[1]{ #1 }
 \newcommand{\MS}[1]{{\color{olive}{}
\fi

\newcommand{\TODO}[1]{\textcolor{red}{#1}}
\newcommand{\R}{\mathbb{R}}
\newcommand{\vv}{\mathbf{v}}
\newcommand{\vp}{\mathbf{p}}
\newcommand{\vs}{\mathbf{s}}

\newcommand{\comment}[1]{}

\newcommand{\argmin}{\operatornamewithlimits{argmin}}

\newcommand{\ConstM}{\emph{ConstantMean}}
\newcommand{\ConstMG}{\emph{GenderMean}}
\newcommand{\ConstMGP}{\emph{GenderPred}}
\newcommand{\IMDB}{\emph{IMDB-100K}}
\newcommand{\IMDBraw}{\emph{IMDB-275K}}
\newcommand{\IMDBportrait}{\emph{IMDB-portrait}}
\newcommand{\IMDBsingle}{\emph{IMDB-23K}} %
\newcommand{\IMDBtest}{\emph{IMDB-test}}
\newcommand{\Labtest}{\emph{Lab-test}}
\newcommand{\PoseNet}{\emph{PoseNet}}
\newcommand{\ShallowNet}{\emph{ShallowNet}}
\newcommand{\DeepNet}{\emph{DeepNet}}
\newcommand{\Linear}{\emph{Linear}}

\begin{abstract}
\noindent Recovering a person's height from a single image is important for virtual garment fitting, autonomous driving and surveillance, however, it
	is also very challenging due to the absence of absolute scale information. We tackle the rarely addressed case, where camera parameters and scene geometry is unknown. %
	To nevertheless resolve the inherent scale ambiguity, we infer height from statistics that are intrinsic to human anatomy and can be estimated from images directly,
	such as articulated pose, bone-length proportions, and facial features.
	Our contribution is twofold.
	First, we experiment with different machine learning models to capture the relation between image content and human height.
	Second, we show that performance is predominantly limited by dataset size and
  create a new dataset that is three magnitudes larger, by mining explicit height labels and propagating them to additional images through face recognition and assignment consistency.
	Our evaluation	shows that monocular height estimation is possible with a MAE
  of 5.56cm.%
\end{abstract}

\section{Introduction}

Estimating people's height from a single image is needed in areas such as subject identification for surveillance purposes, pedestrian distance estimation for autonomous driving, and automated garment fitting in online stores. However, since people's apparent height is affected by camera distance and focal length, assessing someone's real height only from the image is difficult. 

Existing algorithms can only operate under very specific conditions. For example, comparisons to the average height have been used in~\cite{Dey14} to infer people's sizes from group pictures. Similarly, height inference in images acquired with calibrated cameras and in which the location of the ground plane is given~\cite{Guan2009,Zhou2016,JohanVester,Li2011} or where objects of known height are visible~\cite{JohanVester,Ljungberg} have been demonstrated. The method of~\cite{Benabdelkader08} is the only one we know of that can operate on  single uncalibrated RGB images and without prior knowledge. However, it relies on manually supplied keypoints and its reported results on real images lack precision.

In this paper, we propose a more generic and automated approach that relies on
extensive evidence from the biometrics literature that height is correlated to
the relative sizes of body parts
\cite{Adjeroh2010,Zaslan,Shiang1999,Kato1998,Re2013,Mather2010,Burton2013,Wilson2010a,Albanese2016b,Duyar2003},
such as the ratio of the tibia length to the whole body or the head to shoulders
ratio, which are  scale invariant. To this end, we train Deep Nets to capture the correlation between the relative size of body parts without having to explicitly measure them. We demonstrate that they can be trained end-to-end and yield estimates whose mean absolute error is 5.56cm, which is better than the state-of-the-art~\cite{Benabdelkader08}.

Our contribution is empirical in nature with fundamental implications for the
theoretical design of height and pose estimation approaches. First, we have
developed a practical approach to mining a large training dataset via label
propagation. {Second, we show that the specific machine learning algorithm being used
  matters, but only when using training datasets that are several orders of
  magnitude larger than those with only handful of objects, that have been used
  so far~\cite{Ionescu14a,Sigal10,Mehta17a}. Finally, our
  experiments show how important it is to account for small-scale facial details in addition to large-scale full-body information and that these are best captured by networks with scale-specific streams. These findings suggest that future work should focus on both small and large-scale features and must use training datasets that are several magnitudes larger than the ones used currently.

\section{Related Work}
\label{sec:related}

There are several algorithms that can infer age \cite{Rothe16,Malli2016,Dong2016,Wang2015} or emotional state  \cite{Dagar2016,Wegrzyn2017} from single images with high reliability, often exceeding that of humans, in part because these are not affected noticeably by scale ambiguities. By contrast, there are far fewer approaches to estimating human size and we review them briefly here. 

\paragraph{Geometric height estimation.}
The height of standing people can be estimated geometrically from a single image
under some fairly mild assumptions. This can be done by finding head position
and foot contact through triangulation when the camera height and orientation in
relation to the ground plane is known~\cite{Guan2009,Zhou2016,Li2011}, computing
the vanishing point of the ground plane and the height of a reference object in
the scene~\cite{JohanVester,Hartley00}, or accounting for the height of multiple
nearby reference objects \cite{JohanVester,Ljungberg}. However, the necessary knowledge about camera pose, ground plane, and feet contact points is often unavailable.

\paragraph{Height from camera geometry.}
Without external scale information, object size is ambiguous according to the basic pinhole camera model. In practice, lenses have a limited depth of field, which shape-from-defocus techniques exploit~\cite{Mather96,Shi15} to estimate distance. It can be used to guess depth orderings in a single image. However, a focal sweep across multiple images or a specialized camera~\cite{Georgiev13} is required for metric scale reconstruction. 

\paragraph{Height from image features.}
In~\cite{Dey14}, face position and size are used to measure relative heights in pixels first in group pictures and then in image collections featuring groups. Absolute height is estimated from the network of relative heights by enforcing consistency with the average human height, which is effective but only for group photos.  Closest to our approach is the data-driven one of~\cite{Benabdelkader08}, which uses a linear-regressor to predict height from keypoint locations in the input image. The results of an anthropometric survey~\cite{Gordon89} are used to train the regressor. However, even though the keypoints are supplied manually, the results on real images barely exceed what can be done by predicting an average height for all subjects. By contrast, our DeepNet regressor is non-linear, can learn a much more complex mapping that accounts the uncertainty of image-feature extraction, does not require manual annotation of keypoints, and yields better results. 
Its network architecture is inspired by deep networks used for 3D human pose prediction~\cite{Pavlakos16,Tome17,Popa17,Martinez17,Mehta17b,Rogez17,Pavlakos17,Tom2017,Tekin17a}. However, we will show that training on the existing 3D pose datasets with a handful of subjects is insufficient, which was our incentive for creating a larger one.

\paragraph{Height from body measurements.}
Medical studies suggest that the height of an individual can be approximated
given ratios of limb proportions~\cite{Adjeroh2010}, absolute tibia length
\cite{Duyar2003}, foot length~\cite{Zaslan}, and the ratio of head to shoulders
\cite{Shiang1999,Kato1998}. Also human perception of height seems influenced
by head to shoulders ratio, which suggests a real link between head to shoulders ratio to actual height~\cite{Re2013,Mather2010,Burton2013}. There is also a body of anthropological research about inferring
the living height of the individual from the length of several bones in their
skeletons, which indicates that height can be approximated  given the size of some body parts~\cite{Wilson2010a,Albanese2016b,Duyar2003}. 

While these studies indicate that height estimation should be possible from
facial and full-body measurement, there is no easy way to obtain them from
single uncalibrated images and it is not known how naturally occurring feature
extraction error influences accuracy. In particular, the often mentioned absolute length measurements cannot be inferred directly from 2D images.

\section{Method}
\label{sec:method}

Our goal is to estimate human height, $h \in \R$, from a single RGB image, $I \in \R^{3 \times n \times m}$, without prior knowledge of camera geometry, viewpoint position, or ground plane location.
This setting rules out any direct measurement and requires statistical analysis of body proportions and appearance from the images only. We therefore follow a data-driven approach and infer the relationship between image content and human height through machine learning. 

To make the method independent of scene-specific content, we first localize people in the image and then learn a mapping $f_\theta(\bar{I})$ from image crops $\bar{I}$ that tightly contain the target subject.
To this end, we first introduce a diverse dataset of cropped image-height pairs, $D = \{(\bar{I}_i, h_i)\}_i^N$, with $N$ examples (Sec.~\ref{sec:dataset}). Then, we explore different image features and neural network architectures to infer parameters $\theta$ of $f_\theta(I)$ that robustly predict height $h$ given a new input $\bar{I}$ (Sec.~\ref{sec:network}).

\subsection{Dataset Mining}
\label{sec:dataset}

Existing 3D pose datasets are limited to a handful of subjects \cite{Ionescu14a,Sigal10,Mehta17a} (Human3.6Million, HumanEva and MPII-INF-3DHP) and datasets build from web content and comprising anonymous individuals~\cite{Andriluka14,Charles16,Lin14a} (MPII-2D, BBC-Pose, COCO) do not include height information. We therefore built our own. We started from a medium-sized one containing people of known height, which we then enlarged using face-based re-identification and pruned by enforcing label consistency and filtering on 2D pose estimates. Fig.~\ref{fig:datasetExamples} depicts the result.

\begin{figure}
	\centering
	\includegraphics[width=1\linewidth]{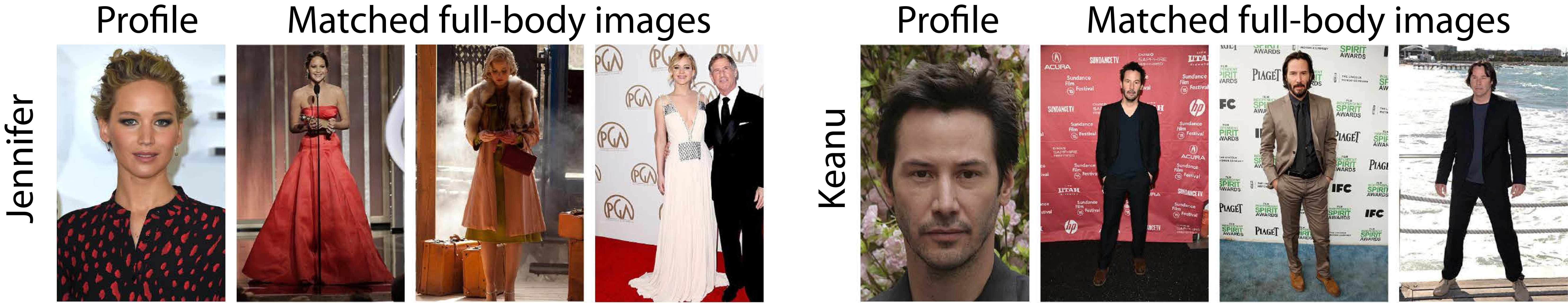}
	\caption{{\bf Examples from \IMDB.} Profile images have been matched to additional images. Thereby, height labels on portrait images are propagated to all the assigned images.}
	\label{fig:datasetExamples}
\end{figure}

\paragraph{Initialization.}

We used the IMDB website as our starting point.  As of February 2018, it covers
8.7 million show-business
personalities.\footnote{\url{https://www.imdb.com/pressroom/stats/}} To find
heights of people  and corresponding images, we crawled the most popular 100.000
actors.\footnote{\url{http://www.imdb.com/search/name?gender=male,female}}   We
found 12,104 individuals with both height information and a profile image
involving a single face.

\vspace{-4mm}
\paragraph{Augmentation.}

IMDB also has more than a million images taken at  award ceremonies and stills
from movies, including full-body images of our 12,104 individuals.  Although
there are associated labels specifying the actors present in the image, these
labels do not specify location of the  person in the image, which makes the
association of height labels to a single person in image potentially ambiguous, especially if there are several people present.

Formally, let $I$ be an image that should be labeled and $S_I$ be the subject labels given by IMDB. We run a face detection algorithm~\cite{dlib09}, which returns a set $K_{I}$ of detected individuals and for each $k \in K_{I}$ the head location in terms of a bounding box and a feature vector $\vv_k$ that describes the appearance compactly.

When there is only one person in the image, we can directly attribute the associated height information to the detected subject.  This has enabled us to create a first annotated dataset of 23,024 examples, which we will refer to as \IMDBsingle.

\begin{figure}
	\centering
	\includegraphics[height=0.22\linewidth]{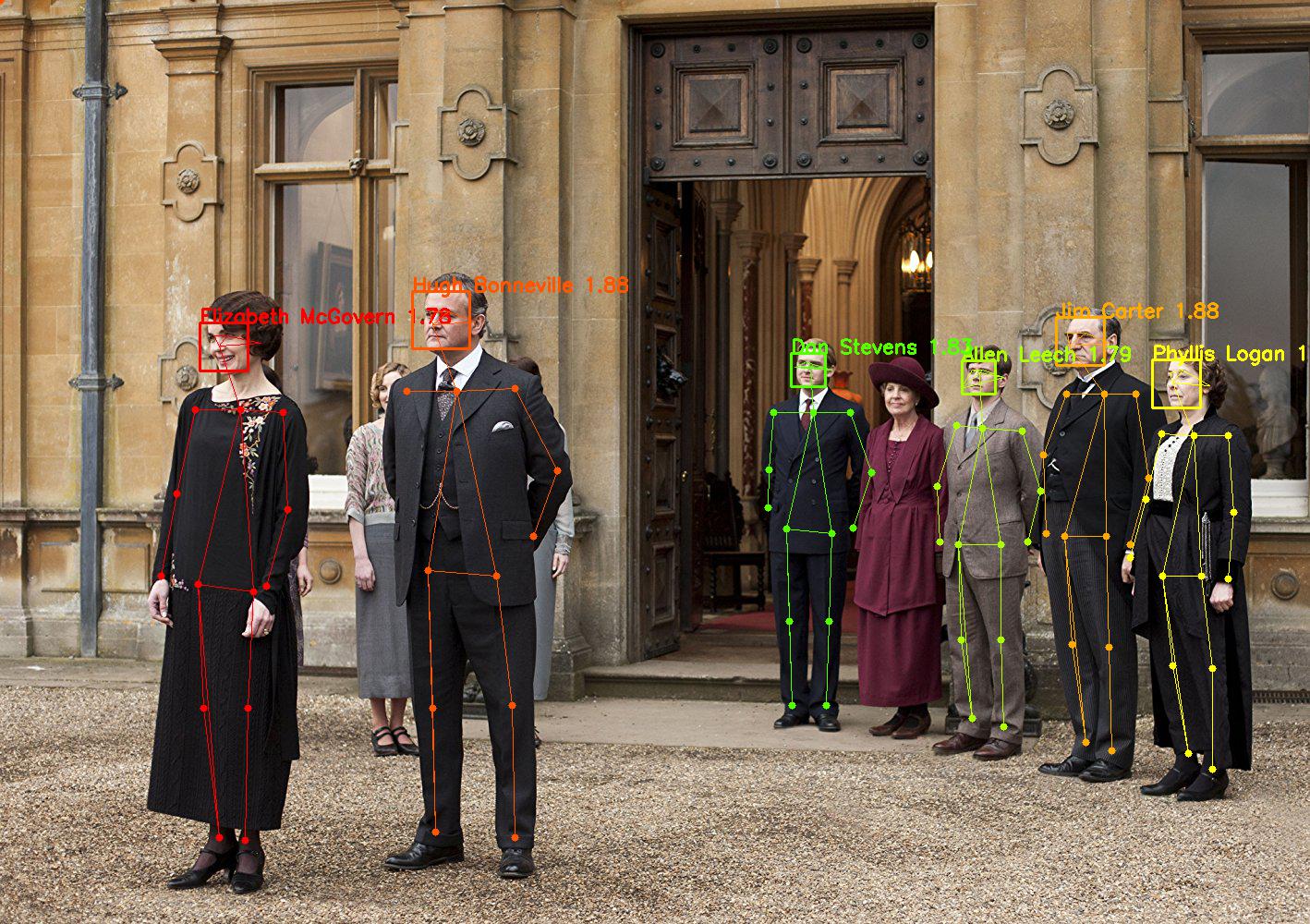}
	\includegraphics[height=0.22\linewidth]{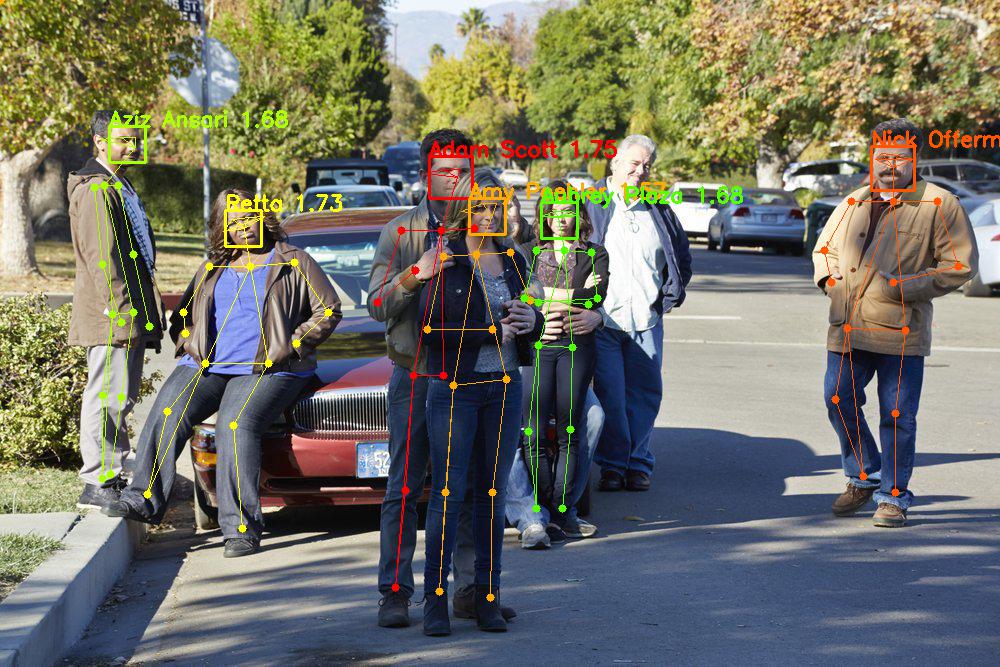}
	\includegraphics[height=0.22\linewidth]{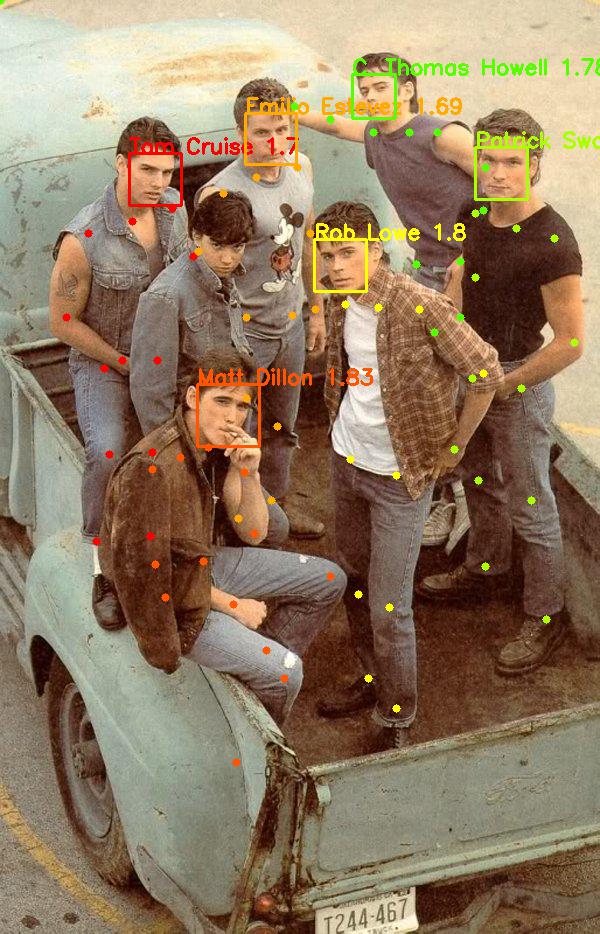}
	\includegraphics[height=0.22\linewidth]{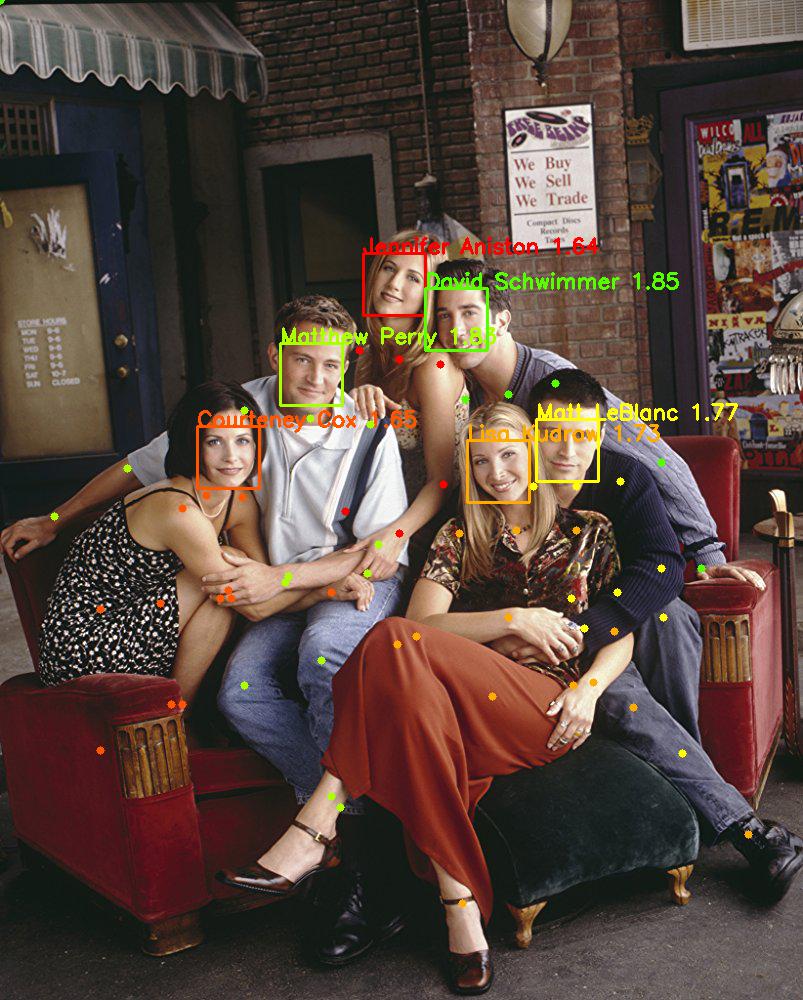}
	\caption{{\bf Identity matching.} Samples from the processed \IMDB{} dataset,
    with an overlay of the assigned subjects' 2D pose, head detection, identity
    and height annotation. In favor of reliable assignments opposed to false assignments,
    some persons remain unassigned.}
	\label{fig:identityMatching}
\end{figure}

The same strategy was used to create the IMDB-WIKI dataset to learn age from facial crops~\cite{Rothe16}. However, we also need the rest of the body and want to create a richer database by also using images with several people and multiple detections.  To associate labels and detections in such cases, we compute from
the profile image of each subject $j \in S_I$ a facial descriptor $\vs_l$ and store its
euclidean distance from comparable descriptors for all detections $\{\vv_k\}_{k \in
  K_{I}}$ in image $I$. This yields a distance matrix $\mathbf{D} \in
\R^{n_K\times n_S}$ between all the $\vv_{k}$ and $\vs_{j}$ descriptors. To
match  one $\vv_{k}$ to a specific $\vs_{j}$, we make sure that 
$\mathbf{D}[k,j]$ is smaller than all other distances in row $k$ and column $j$,
that is $\vv_{k}$ is the closest match to all $\vs$ in the list of subjects and,
similarly, $\vs_{j}$ is the closest match to all $\vv$ in the list of
detections. In practice, we apply an additional ratio test to ensure the assignment is reliable:
We assign $\vs_{j}$ to the best matching feature vector $k^* = \argmin_k
\mathbf{D}[k,j]$, but only if the quotient $q = \frac{\mathbf{D}[k^*,j]}{
  \min_{k \neq k^*} \mathbf{D}[k,j]}$ is smaller than $\tau=0.9$.
That means best match must be significantly better than the second-best
match to be accepted.
This produces a much
larger set of 274,964 image-person pairs with known height, which we will refer
to as \IMDBraw.  We show a few examples in Fig.~\ref{fig:identityMatching}.

To estimate the accuracy of our assingments in \IMDBraw we randomly select 120 images which
include multiple faces, where assigning identity to faces is non-trivial and possibly
 erroneous. Out of the 331 IMDB labels in 120 images, we assigned
237 labels to faces inside the images, where only 5 of the assignments were
wrong. We also repeated the same experiment for \IMDBsingle, where assignment is
much easier. We again selected 120
random images and check the accuracy of the assignments. We observed only a
single mismatch. Overall this corresponds to an
estimated label precision of 98.0\% and recall of 70.1\%. 

\vspace{-4mm}
\paragraph{Filtering and preprocessing.} 

As discussed in the Related Work section, previous studies suggest that full-body pose and bone-length relations contain scale information. Therefore, we run a multi-person 2D pose estimation algorithm~\cite{Cao17} on each dataset image $I$ and assign the detected joints to the subject whose estimated head location as predicted by the face detector~\cite{dlib09} is closest to the one estimated by the 2D pose extraction algorithm.  The 2D joints are then used to compute image crops $\bar{I}$ that tightly enclose the body and head. The face is similarly cropped to $\tilde{I}$, as shown in the left side of Fig.~\ref{fig:network_architecture}. 

Finally, we automatically exclude from \IMDBraw{}  images missing upper body joints or whose crop is less than 32 pixels tall. This leaves us with 101,664 examples, which we will refer to as the \IMDB{} dataset. We also applied this process to \IMDBsingle{}. In both cases, we store for each person the annotated height $h$, a face crop $\tilde{I}$, a facial feature vector $\vv$, a pose crop $\bar{I}$, a set of 2D joint locations $\vp$, and  the gender if available.

\vspace{-4mm}
\paragraph{Splitting the dataset.}
We split \IMDB{} into three sets, roughly in size 80k, 15k and 5k
images for training, testing and validaton, respectively.

\subsection{Height Regression}
\label{sec:network}

Since there is very little prior work on estimating human height directly from image features, it is unclear which features are the most effective. We therefore tested a wide range of them. To the face and body crops, $\tilde{I}$ and $\bar{I}$, discussed in Section~\ref{sec:dataset}, which we padded to be $256\times256$, we added the corresponding 2D body poses, in the form of 2D locations of keypoints centered around their mean and whitened, along with 4096-dimensional facial features computed from the last hidden layer of the VGG-16-based face recognition network of \cite{Wu}.

\begin{figure}
	\centering
	\includegraphics[width=1\linewidth]{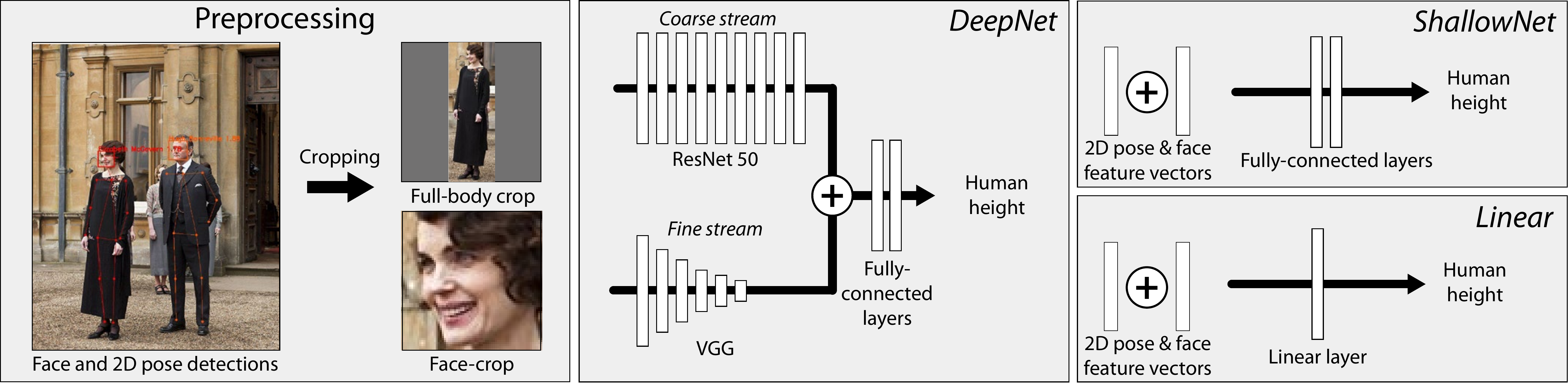}%
	\caption{{\bf Deep two-stream architecture.} Humans are automatically detected
    and cropped in the preprocessing state. We experiment with \DeepNet, a
    two-scale deep convolutional network that is trained end-to-end, and a
    simple \ShallowNet, that operates on generic image features, and with
    \Linear, which is a simple Linear Regression.
    }
	\label{fig:network_architecture}
\end{figure}

Given all these features, we tested the three different approaches to regression depicted by Fig.~\ref{fig:network_architecture}: 
\begin{itemize}

 \item \Linear{}. Linear regression from the pre-computed 2D pose and facial features vectors, as in~\cite{Benabdelkader08}. 
 
 \item \ShallowNet{}. Regression using a 4-layer fully connected network as used
   in \cite{Martinez17}. \ShallowNet{} operates on the same features as \Linear{}.
 
 \item \DeepNet{}. Regression using a deeper and more complex network to combine
   fine-grained facial features with large-scale information about overall body
   pose and shape. It uses two separate channels to compute face and full body
   features directly from the body and face crops, respectively, and uses two
   fully connected layers to fuse the results, as depicted in
   Fig.~\ref{fig:network_architecture}. By contrast to {\bf ShallowNet}, we
   train this network end-to-end and thereby optimize the facial and full-body
   feature extraction networks for the task of human height estimation using MSE
   Loss. To allow for a fair comparison, we use the same VGG architecture in the face stream\cite{Wu}. For the full body one, we utilize a ResNet~\cite{He16}.

\end{itemize}

\section{Evaluation}

We now quantify the improvement brought about by our estimation and try to
tease out the influence of its individual components, dataset mining, and
network design. We also show some example results on ~Fig.\ref{fig:labtest}
for the most popular actors from the test split of \IMDB.

\paragraph{{\bf Metrics.}} We report height estimation accuracy in terms of the mean absolute error (MAE) compared to the annotated height in cm.  We also supply cumulative error histograms.

\begin{figure*}
\begingroup
\setlength{\tabcolsep}{0.5pt}
\renewcommand{\arraystretch}{0.7}
\begin{center}
  \begin{tabular}{ccccccccc}
\includegraphics[width=0.1\linewidth]{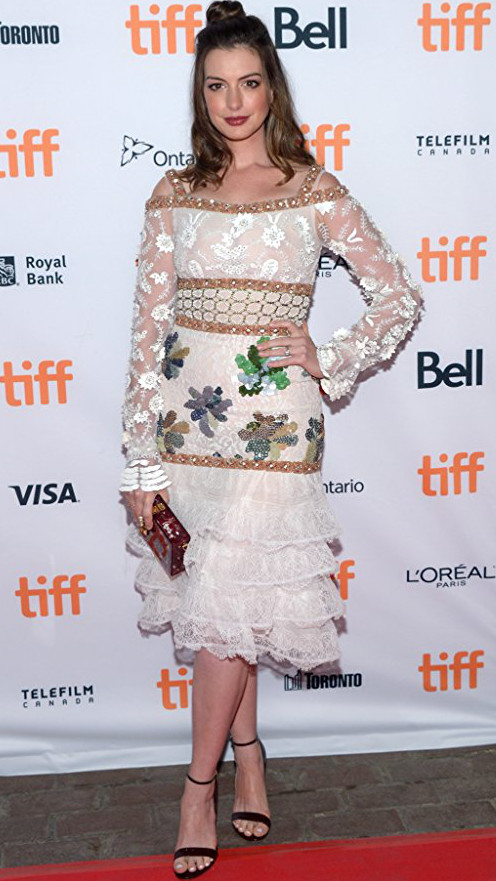} & \includegraphics[width=0.1\linewidth]{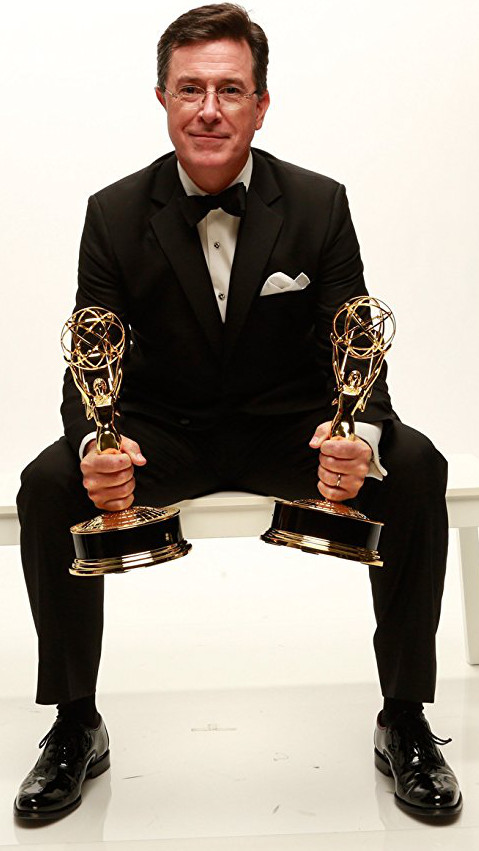} & \includegraphics[width=0.1\linewidth]{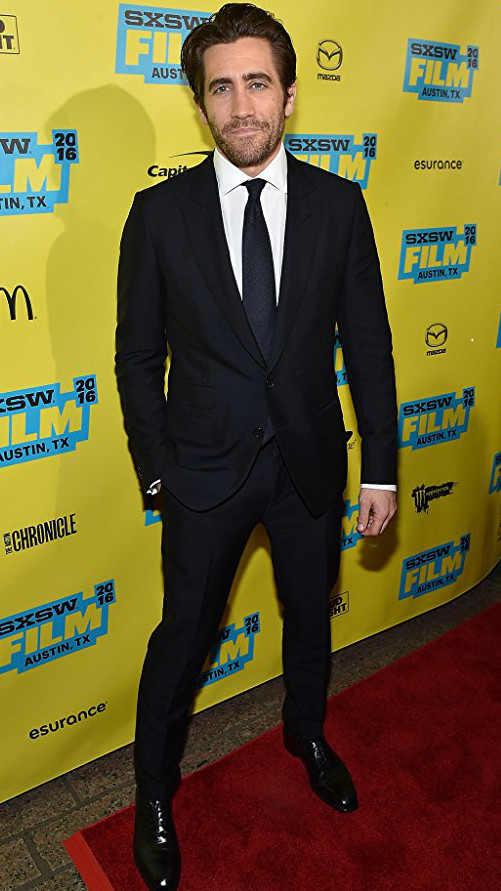} & \includegraphics[width=0.1\linewidth]{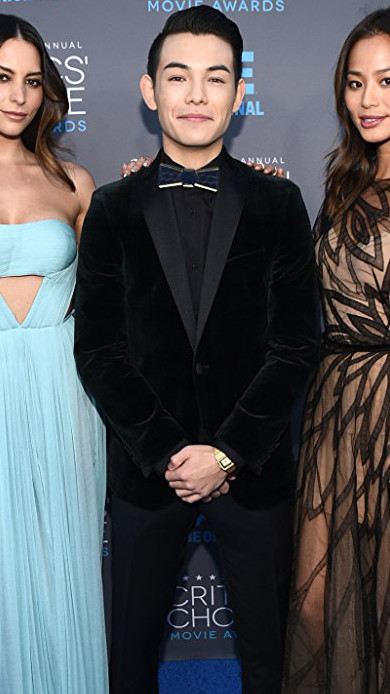} &\includegraphics[width=0.1\linewidth]{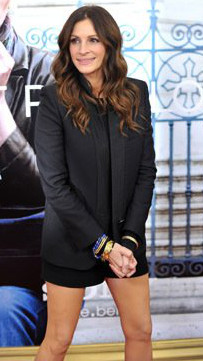} & \includegraphics[width=0.1\linewidth]{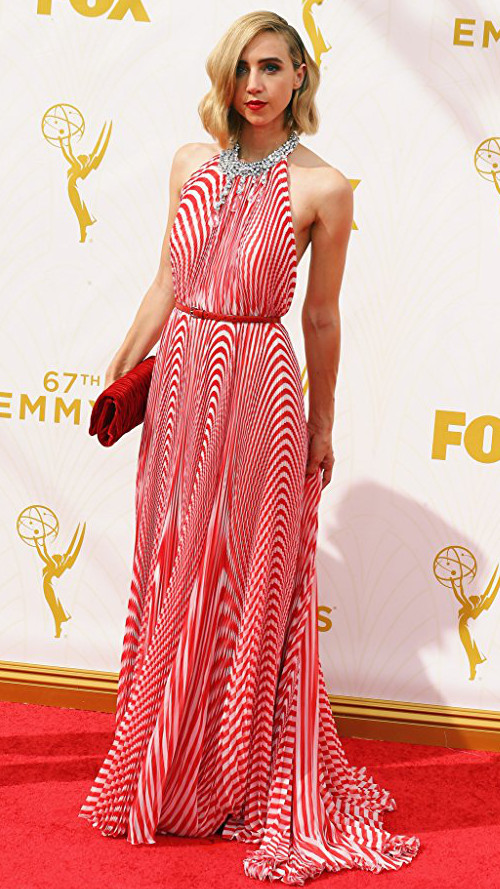} & \includegraphics[width=0.1\linewidth]{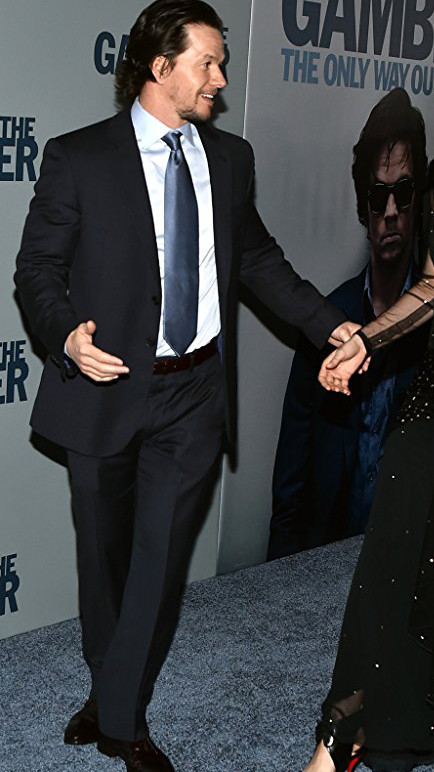} & \includegraphics[width=0.1\linewidth]{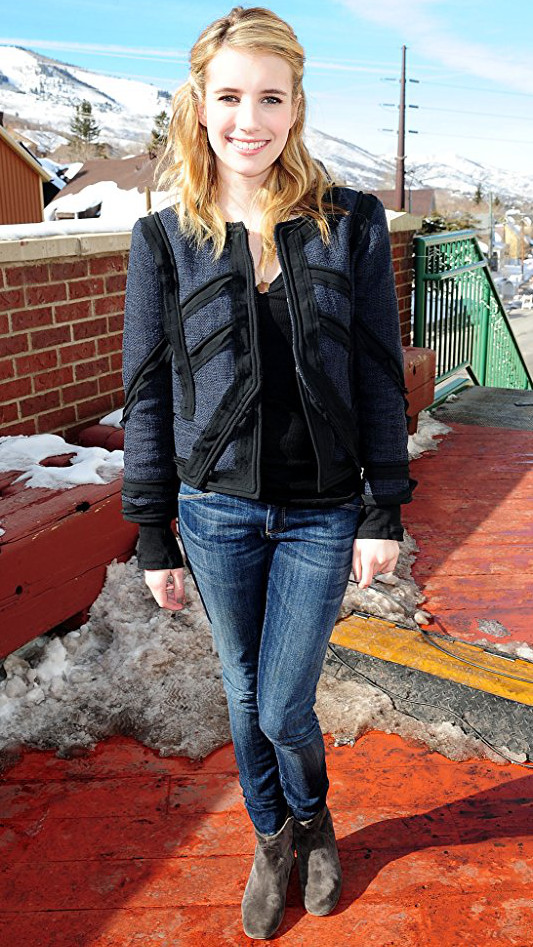} & \includegraphics[width=0.1\linewidth]{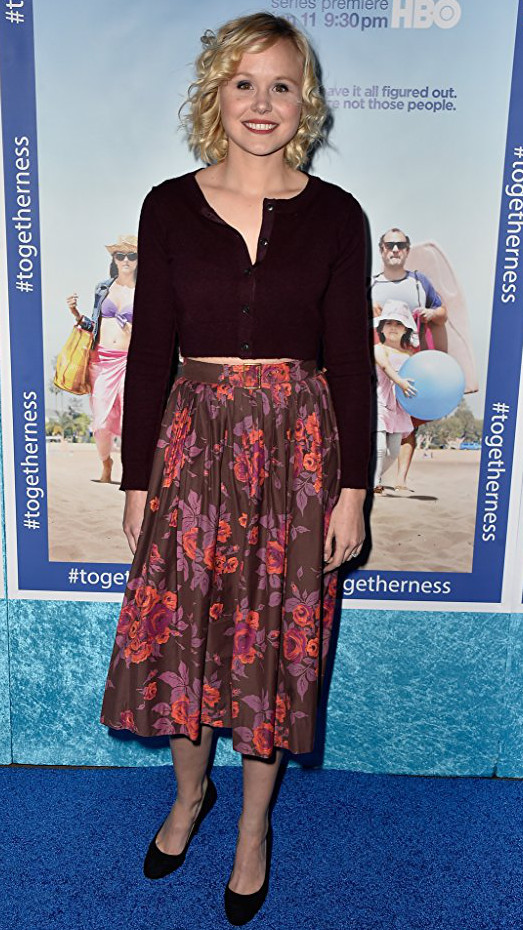} \\
  166/170 & 181/179 & 181/181 & 177/173 & 164/173 & 166/164 & 182/173 & 163/156 & 167/168  
\end{tabular}
\end{center}
\endgroup
	  	\caption{{\bf Qualitative evaluation.} Results on test set of \IMDBtest{} shown as prediction/ground-truth in centimeters.} 
    
	\label{fig:labtest}
\end{figure*}

\noindent{\bf Independent test set.}
To demonstrate that our training dataset is generic and that our models
generalize well, we created \Labtest, an in-house dataset containing photos of
various subjects whose height is known precisely.  Since it was acquired completely independently from IMDB, we can be sure  that our results are not contaminated by overfitting to specific poses, appearance, illumination, or angle consistency. \Labtest~ depicts 14 different individuals with 10 photos each. Each one contains a full body shot in different settings, sometimes with small occlusions to reflect the complexities of the real world. The subjects are walking in different directions, standing, or sitting. Individuals span diverse ethnicities from several European and Asian countries and heights, ranging from 1.57 to 1.93 in meters. 

\noindent{\bf Baselines.}
We compare against the following baselines in order of increasing sophistication:
\begin{itemize}

 \vspace{-1mm}
 \item  \ConstM. The simplest we can do, which is to directly predict  the average height of
   of \IMDB{}, which is 170.1 centimeters.
 
 \vspace{-1mm}
 \item \ConstMG. Since men are taller than women on average, gender is a predictor of height. We use the ground-truth  annotation as an oracle and the gender-specific mean height as the prediction. 
 
 \vspace{-1mm}
 \item  \ConstMGP. Instead of using an oracle, we train a network whose architecture is similar to \DeepNet{} to predict gender instead of height and again use the gender-specific mean height as the prediction. 

 \vspace{-1mm}
 \item  \PoseNet. We re-implemented the method of~\cite{Mehta17a} that predicts 3D human pose in absolute metric coordinates after training on the Human3.6M dataset \cite{Ionescu14a}. Height information is extracted from the predicted bone lengths from head to ankle. To accommodate for the distance from ankle to the ground, we find a constant offset between the predicted height and the ground truth height on \IMDB. 

\end{itemize}

\subsection{Comparative Results}

\begin{table}[b]
\center
\resizebox{0.70\width}{!}
{\begin{tabular}{cc}
   \begin{tabular}{|l|c|c|c|c|c|c|}
		\hline
		& \multicolumn{3}{c|}{$\IMDB$} & \multicolumn{1}{c|}{$\Labtest$}\\
		Method & all & women & men & all  \\
		\hline
		\ConstM & 8.25  & 7.46 & 9.22 & 11.0 \\
		\ConstMGP &  6.61 & 6.28 & 7.12 & 9.26 \\
		\PoseNet~\cite{Mehta17a} & - & - & - & 10.65\\
		\DeepNet~(ours) & \bf 6.14 & \bf 5.88 & \bf 6.40 & \bf 9.13 \\
		\hline
		\ConstMG & 5.91 & 5.63 & 6.23 & 8.66 \\
		\DeepNet~(gender-specific) & \bf 5.56 & \bf 5.23 & \bf 6.03  & \bf 8.53 \\
		\hline
   \end{tabular}
   &
   \begin{tabular}{|l|c|c|c|}
			\hline
			& \multicolumn{3}{c|}{Regression type} \\
			Input features & Linear & \ShallowNet & \DeepNet\\
			\hline
                         Body crop only & 7.56 / 11.10 & 7.10 / 10.40 & 6.40 / 9.43 \\
			\hline
			Face crop only & 6.49 / 10.25 &  6.31 / 9.99 & 6.25 / {\bf 8.87}  \\
			\hline
                         Body and Face  & 6.40 / 10.2 &  6.29 / 9.92 & {\bf 6.06} / 9.13  \\
			\hline
    \end{tabular}\\
    (a)&(b)
\end{tabular}}
\caption{\small {\bf Mean Absolute Error (MAE) on \IMDB{} and \Labtest{}.} (a)
  Comparison against our baselines. (b) Ablation study, accuracies are given in
  \IMDB{} / \Labtest{} format. }
\label{tab:accuracy}
\end{table}

We report our mean accuracies on \IMDB and  \Labtest{} along with those of the baselines at the top Tab.~\ref{tab:accuracy}(a). \DeepNet{}, which is our complete approach, outperforms them  on both, with  \ConstMGP{} being a close second. This confirms that gender is indeed a strong height predictor. 
To confirm this, we retrained \DeepNet{} for men and women separately and compare its accuracy to that of $\ConstMG$ at the bottom of the table. As can be seen, our full approach improves upon this as well but, somewhat unexpectedly, more for women than men. 

\PoseNet{} does not do particularly well, presumably because it has been trained on many images but all from a small number of subjects. It has therefore not learned the vast variety of possible body shapes. 

In Table~\ref{tab:accuracy}(b), we report the results of an ablation study in which we ran the three versions of our algorithm---\Linear{}, \ShallowNet{}, and \DeepNet{} introduced in Section~\ref{sec:network}---on the full dataset, on the faces only, or on the body only. In all cases, \DeepNet{} does better than the others, which indicates that it also outperforms the state-of-the-art algorithm~\cite{Benabdelkader08}, which \Linear{} emulates. 

On \IMDB{}, using both body and faces helps but, surprisingly, not on \Labtest{}
where using the faces only is the best.
We suspect that the poses in  \Labtest{} are more varied than in \IMDB{}. 
Furthermore, there is also wider spread of heights in \Labtest{} and other biases due to its small size, which might contribute to this unexpected behavior. 
This is something we will have to investigate further.  Aside from this, the conclusions drawn from experiments on \IMDB{} are all confirmed on \Labtest{}.

\begin{figure}
	\centering
	\includegraphics[height=0.4\linewidth]{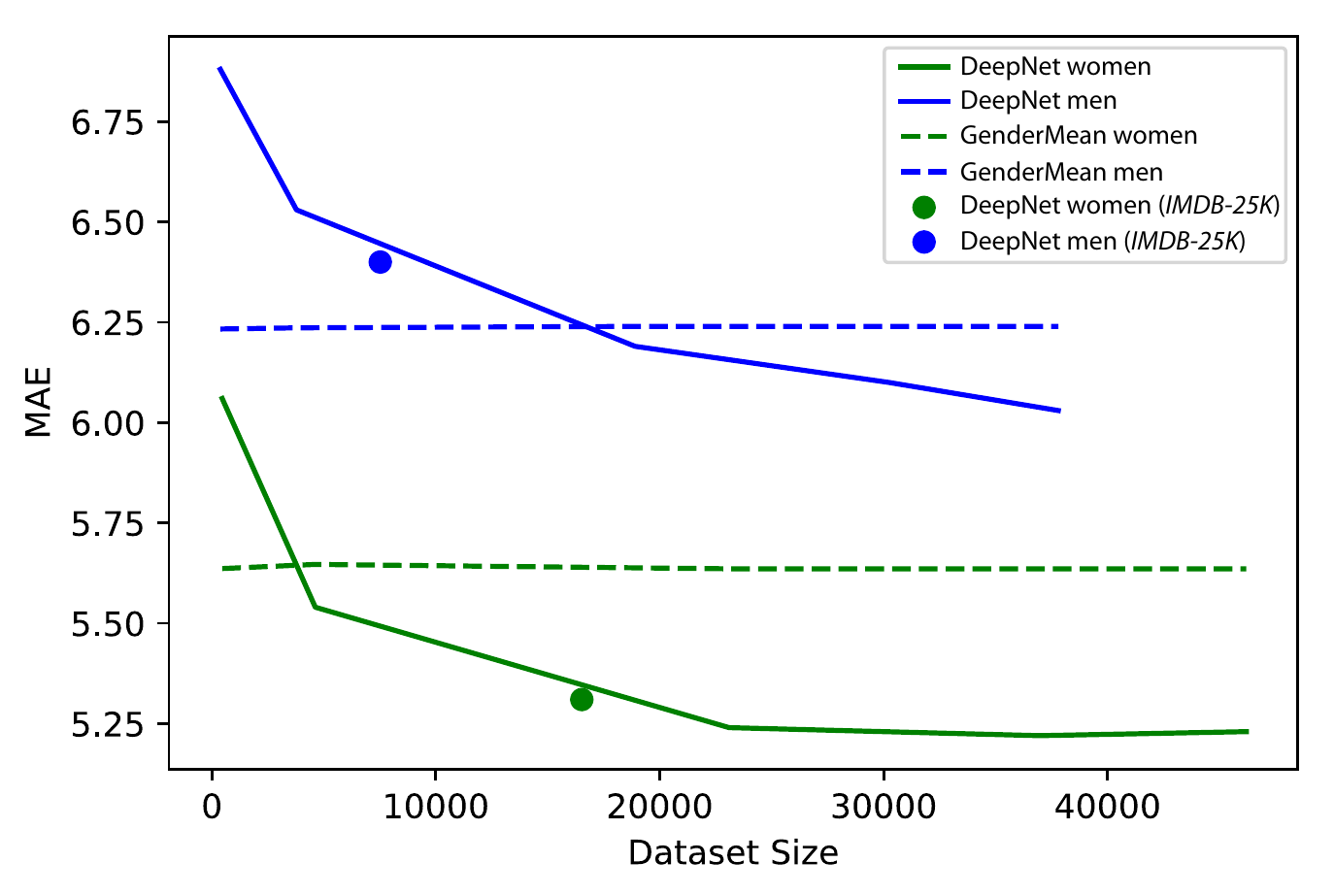}
	\caption{{\bf Accuracy as a function of the training dataset size.} We plot separate curves for men and women. }
	\label{fig:datasetSize}
\end{figure}

In Fig.~\ref{fig:datasetSize}, we plot the accuracy of our model as a function
of the size of the training set. It clearly takes more than 10,000 to 20,000
images to outperform \ConstMG{}. Interestingly, it seems to take more images for
men than women, possibly due to the larger variance in men height. This indicates the results we report here for men might not be optimal yet and would benefit from using an even larger training set.
This finding could also explain the lower accuracy on \Labtest, which is dominated by men, compared to \IMDB.

\section{Discussion and Limitations}

Although we collected a large set of images and actors with height annotations, our dataset comes with caveats: Height information given in IMDB might be imprecise or even speculative and there is no practical way to assess the quality of the annotations. Furthermore, human height can change with age and our dataset only provides a single number, even if someone's images span multiple decades. Therefore we assume the provided height information is descriptive rather than precise. One future direction of research can be eliminating such inaccuracies, possibly by making annotation consistent in group pictures \cite{Dey14}. Furthermore, if some annotations can be identified as unreliable, this could be modeled by incorporating a confidence value during training and prediction. %

\section{Conclusion}

With 274,964 images and 12,104 actors, we introduce the largest
dataset for height estimation to date. The provided label association could be used not only for height estimation but also to explore other properties of human appearance and shape in the future. We experimented with different network architectures that, when trained on IMDB-100K, improves significantly on existing height estimation solutions. 

Our findings have several implications on future work. Since height prediction from a single image remains inaccurate, methods for 3D human pose predictions should not be evaluated in metric space, as is often done, but after scale normalization, and the inevitable inaccuracies in height estimation evaluated separately. Furthermore, if absolute height is desired, a large dataset must be used to cover the large variations in human shape, pose and appearance. Finally, it is important to combine both facial and full-body information for height regression.
\bibliographystyle{splncs}
\bibliography{../../../../bibtex/string,../../../../bibtex/vision,../../../../bibtex/learning,../../../../bibtex/misc,../../../../bibtex/biomed,egbib}

\end{document}